\documentclass{article}
\usepackage[T1]{fontenc} 
\usepackage[utf8]{inputenc} 
\usepackage{ismir,amsmath,cite,url}
\usepackage{graphicx}
\usepackage{color}
\usepackage{multirow}
\usepackage{adjustbox}
\usepackage{array}
\usepackage{makecell}
\usepackage{color,soul}
\usepackage{kotex}
\usepackage{booktabs}
\usepackage{subfigure}
\usepackage{comment}
\usepackage[dvipsnames]{xcolor}
\usepackage{tabularx}
\usepackage{array}
\newcolumntype{L}{>{\centering\arraybackslash}m{1cm}}
\newcolumntype{R}{>{\centering\arraybackslash}m{5cm}}

\title{Zero-shot Learning for Audio-based Music Classification and Tagging}

\multauthor
{Jeong Choi$^1$\textsuperscript{*} \hspace{1cm} Jongpil Lee$^1$\textsuperscript{*} \hspace{1cm} Jiyoung Park$^2$ \hspace{1cm} Juhan Nam$^1$} 
{ 
$^1$ Graduate School of Culture Technology, KAIST\\
$^2$ NAVER Corp.\\
{\tt\small  \{jeong.choi, richter, juhannam\}@kaist.ac.kr, j.y.park@navercorp.com}
\\
\small{* Equally contributing authors.}
}

\def\authorname{Jeong Choi, Jongpil Lee, Jiyoung Park, Juhan Nam}

\begin{document}

\maketitle

\begin{abstract}
Audio-based music classification and tagging is typically based on categorical supervised learning with a fixed set of labels. This intrinsically cannot handle unseen labels such as newly added music genres or semantic words that users arbitrarily choose for music retrieval.  
Zero-shot learning can address this problem by leveraging an additional semantic space of labels where side information about the labels is used to unveil the relationship between each other. 
In this work, we investigate the zero-shot learning in the music domain and organize two different setups of side information. 
One is using human-labeled attribute information based on Free Music Archive and OpenMIC-2018 datasets. The other is using general word semantic information based on Million Song Dataset and Last.fm tag annotations. 
Considering a music track is usually multi-labeled in music classification and tagging datasets, we also propose a data split scheme and associated evaluation settings for the multi-label zero-shot learning. 
Finally, we report experimental results and discuss the effectiveness and new possibilities of zero-shot learning in the music domain.

\end{abstract}
\section{Introduction}\label{sec:introduction}
Audio-based music classification and tagging is a task that predicts musical categories or attributes such as genre, mood, instruments and other song quality from music tracks. Current state-of-the-arts algorithms are based on supervised learning of deep convolutional neural networks that directly predict the labels in the output layer \cite{nam2019}. That is, the neural networks are trained to minimize the prediction errors with regards to the labels. The prediction results can be used to automatically annotate music tracks or retrieve music tracks using the labels as a query word \cite{turnbull2008semantic}. By the nature of the setting in the supervised learning, however, the approach allows only a fixed set of word labels in the annotation and retrieval.

\begin{figure*}
\centering\includegraphics[width=0.9\linewidth]{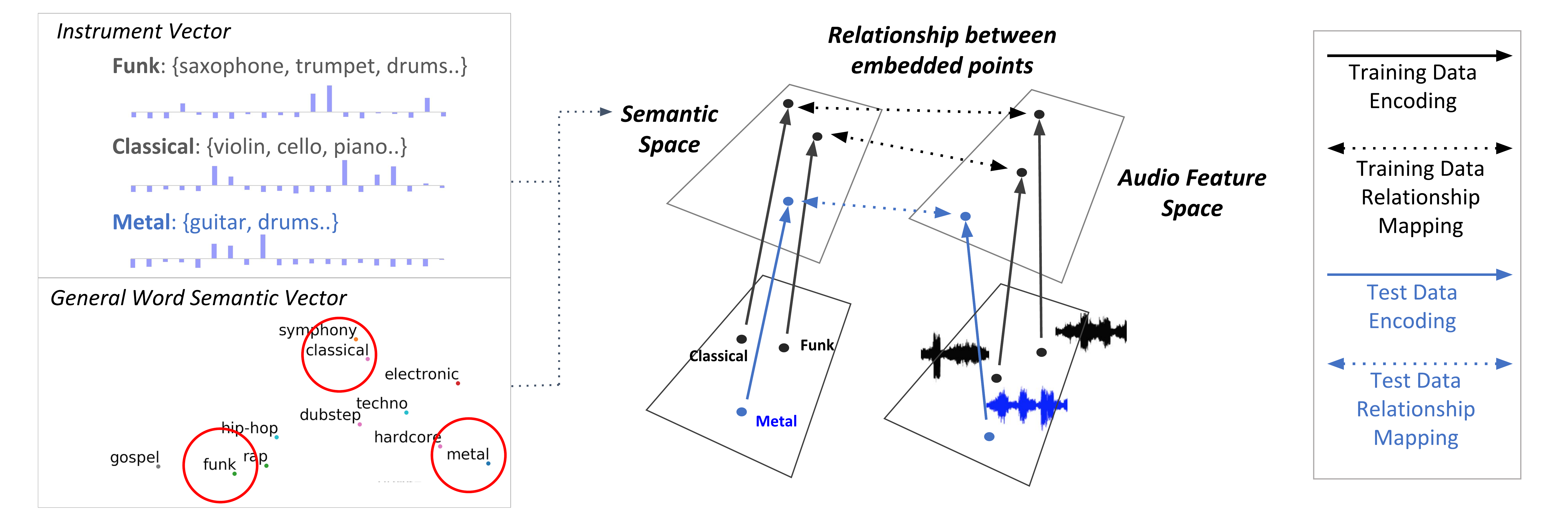}
\caption{Overview of zero-shot learning concept applied to music classification and tagging.}
\label{fig:figure1}
\end{figure*}

Zero-shot learning is a learning paradigm that can overcome this limitation and enables the trained model to predict unseen labels \cite{palatucci2009zero,awa}. For example, it allows the model to predict newly added music genres after the training or retrieve songs using a query word that users arbitrarily choose. This is possible by utilizing side information that derives a separate semantic space from labels. For example, the side information can be musical instrument annotation vectors of music genres or word embedding learned from sentences. The zero-shot learning approach conducts supervised learning between the semantic space and audio feature space. Once the mapping between two embedding spaces is learned, the model can predict unseen labels. Figure \ref{fig:figure1} illustrates the concept of zero-shot learning applied to music classification and tagging.

The zero-shot learning approach was previously applied to music data \cite{sandouk2016multi}. However, they focused on evaluating a specific semantic embedding method that works for general multimedia data rather than delving into zero-shot learning in the music domain. In this work, we carefully investigate how the concept of zero-shot learning can be properly applied to audio-based music classification and tagging. Specifically, we designed two settings of side information. One is using human-labeled attribute information and the other is using general word semantic information.
Also, considering a  music track is usually  multi-labeled in music classification and tagging datasets, we propose a data split scheme that yields a comprehensive list of combinations for seen or unseen audio and labels, and evaluate them in the various settings. Through the experiments, we show the effectiveness and new possibilities of zero-shot learning in the music domain.   



\section{Background: Zero-Shot Learning} \label{sec:section2}

Zero-shot learning has been studied mainly for object recognition in the field of computer vision \cite{wang2019survey}. They have attempted to build a model that can recognize novel visual categories without any associated training samples by employing a joint embedding space of both images and their class labels \cite{wang2019survey}. It was originally inspired by human's ability to recognize objects without seeing training examples or even create new categories dynamically based on semantic analysis \cite{fu2017recent}. What enables this semantic exploitation of unseen images is ``information transfer'' that applies the knowledge learned in an auxiliary domain to the main targeted task. Thus, it aims to explore how seen and unseen images are semantically related based on the side information.

The types of side information can be largely divided into two categories \cite{wang2019survey,fu2017recent,gzsl}. One is human-annotated attributes of the labels. In the computer vision community, there are publicly available human-annotated attribute datasets such as AWA (Animals With Attributes)\cite{awa} and CUB (Caltech-UCSD Birds-200)\cite{welinder2010caltech}. For example, AWA offers images of 50 animals annotated on 85 attributes associated with animals characteristics such as ``furry'' or ``has tail''. These examples may be equivalent to music genre classification where corresponding attributes are musical instruments. These attributes can be used as binary output to train a classifier and infer unseen class based on similarity measure \cite{awa,al2016recovering} or to learn their relative strength \cite{parikh2011relative,singh2016end}. When such explicit attributes data are not available, the semantic attributes can be learned with hierarchy of classes or other relationships \cite{costa,rohrbach2010whathelps,reed2016learning}. In general, this attribute-based approach has advantages in interpretability but requires well-defined attributes and annotation data. 

The other type of side information is a general word semantic space learned from different resources. A commonly used choice is neural language models such as Word2Vec \cite{word2vec, skipgram, akata2015SJE, xian2016latent, devise, conse} and GloVe \cite{glove, akata2015SJE, xian2016latent}. Another choice is learning relational semantic embeddings using pre-defined lexical hierarchy such as WordNet \cite{wordnet,rohrbach2010whathelps}.  These general semantic spaces have an advantage in that they have a large set of words in the vocabulary to predict unseen labels. However, if the target labels have a specific context, the general semantic space may fail to capture it \cite{sandouk2016multi}.

\section{Data Split Scheme for Multi-Label Zero-Shot Learning}
\label{sec:section2.2}

Many of music classification and tagging datasets have multiple labels to annotate music tracks. However, zero-shot learning in the image domain has been treated primarily as a single-label problem, although a few studies have attempted to address it in multi-label classification \cite{costa,fu2015transductive}. An important difference between single-label and multi-label zero-shot learning is data split scheme for training and test. In general, in zero-shot learning, both data and labels are split into seen and unseen sets. In the single-label setup, the label split can automatically divide the dataset into training and test sets (Figure \ref{fig:figure2} (a)). In the multi-label setup, however, specifying reasonable instance or label splits is not straightforward.

\subsection{Previous Approaches}

\begin{figure}[ht]
\centering\includegraphics[width=0.8\columnwidth]{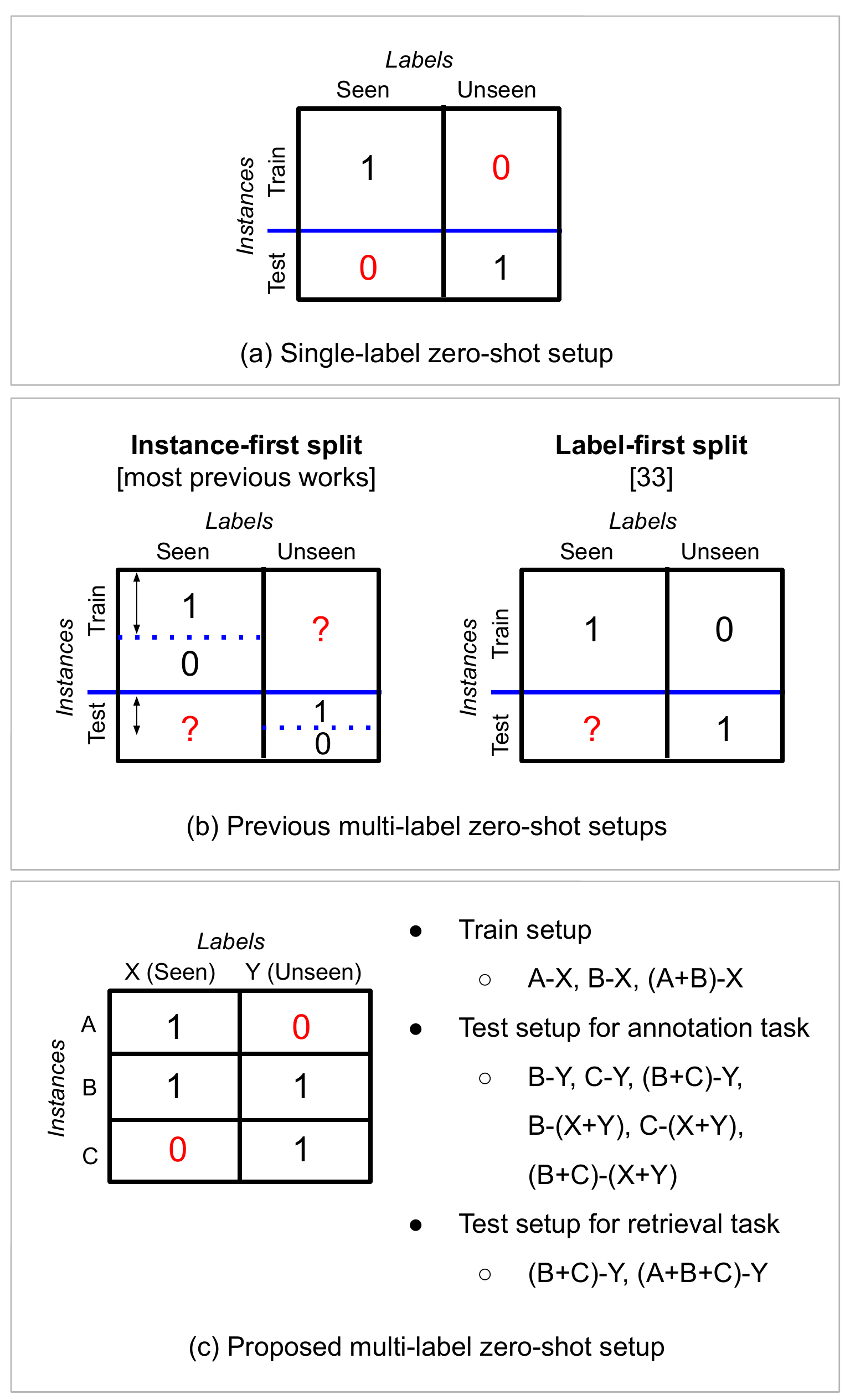}
\vspace{-2mm}
\caption{Data split for zero-shot learning }
\label{fig:figure2}
\end{figure}

Most of previous works on multi-label zero-shot learning are conducted the instance-first split \cite{ren2015multi}. They first split instances into train and test, and only used seen labels for training as shown in the left of Figure \ref{fig:figure2} (b). In this case, some of the instances in the train set can have positive annotations for unseen labels. As an alternative, the label-first split was proposed in \cite{wang2017multi}. They first split labels into seen and unseen groups, and select training instances to have no positive annotation for unseen labels and select test instances to have at least positive annotation on unseen labels as shown in the right of Figure \ref{fig:figure2} (b). However, in this case, due to the nature of multi-label data, too many instances can be assigned to the test set.

Meanwhile, it is an important issue to determine which split in instance (train and test) or label (seen and unseen) should be evaluated in measuring zero-shot learning performance. A generalized zero-shot learning evaluation setting is proposed in \cite{gzsl}. It includes both seen and unseen labels at test time to examine more natural annotation performance compared to use only unseen labels at test time. In multi-label zero-shot learning, however, the data split and evaluation settings are still not clear and there is no agreed consensus yet.

\subsection{Proposed Data Split Scheme}
\label{sec:proposed_data_split_scheme}

We propose a data split scheme and evaluation settings for multi-label zero-shot learning to measure the performance in more refined and various settings. The proposed data split is shown in Figure \ref{fig:figure2} (c). We first divide labels into seen (X) and unseen (Y) groups and then split instances into three groups. The first subset (A) of instances are labeled with at least one from seen labels and not labeled with any of unseen labels. The second subset (B) of instances are labeled with at least one from each of seen and unseen labels. Lastly, the third subset (C) of instances are only labeled with at least one from unseen labels. 
Then we create the three setups (including train setup, test setup for annotation task, and test setup for retrieval task) with a combination of instance subsets (A, B, and C) and label subsets (X and Y) as described in Figure \ref{fig:figure2} (c).

In this configuration, the label-first split can be obtained when the training set is A-X (a subset where instances are from A and labels are from X) and the test set is (B+C)-Y (a subset where instances are from (B+C) and labels are from Y). Also, the generalized zero-shot learning evaluation setting can be expressed to include not only the Y area but also the (X+Y) area at test time. We therefore extend the split and evaluation settings more comprehensively, allowing for many aspects of multi-label zero-shot learning to be considered.

We take the train setup from A-X, B-X, and (A+B)-X. The instances in A-X only contain annotations on seen labels and the instances in B-X contain annotations on both seen and unseen labels. By distinguishing these two areas, we expect to see the difference in model learning by using instances with and without annotations in the unseen labels in multi-label zero-shot learning.

The test setup for the annotation task is made up of combinations of B, C, or (B+C) with Y. In addition, combinations of B, C, or (B+C) with (X+Y) can be also considered to measure generalized zero-shot learning performance on the annotation task as explained in the Section \ref{sec:section2.2}. The test setup for retrieval task is composed of (B+C)-Y and (A+B+C)-Y. We can regard (A+B+C)-Y as a case of generalized zero-shot learning evaluation setting because the retrieval is performed not only on the instances of unseen labels ((B+C) split) but also on instances of seen labels (A split).
The C-Y area cannot be formed in the retrieval evaluation. The reason for this is that once we split labels into seen and unseen and then assign overlapping instances to the B area, we may not guarantee that all the unseen labels have at least one positive activation on the instances in the C area.

\section{Model}


\subsection{Deep Embedding Model}

\begin{figure}[!t]
\vspace{-2mm}
\centering\includegraphics[width=0.96\columnwidth]{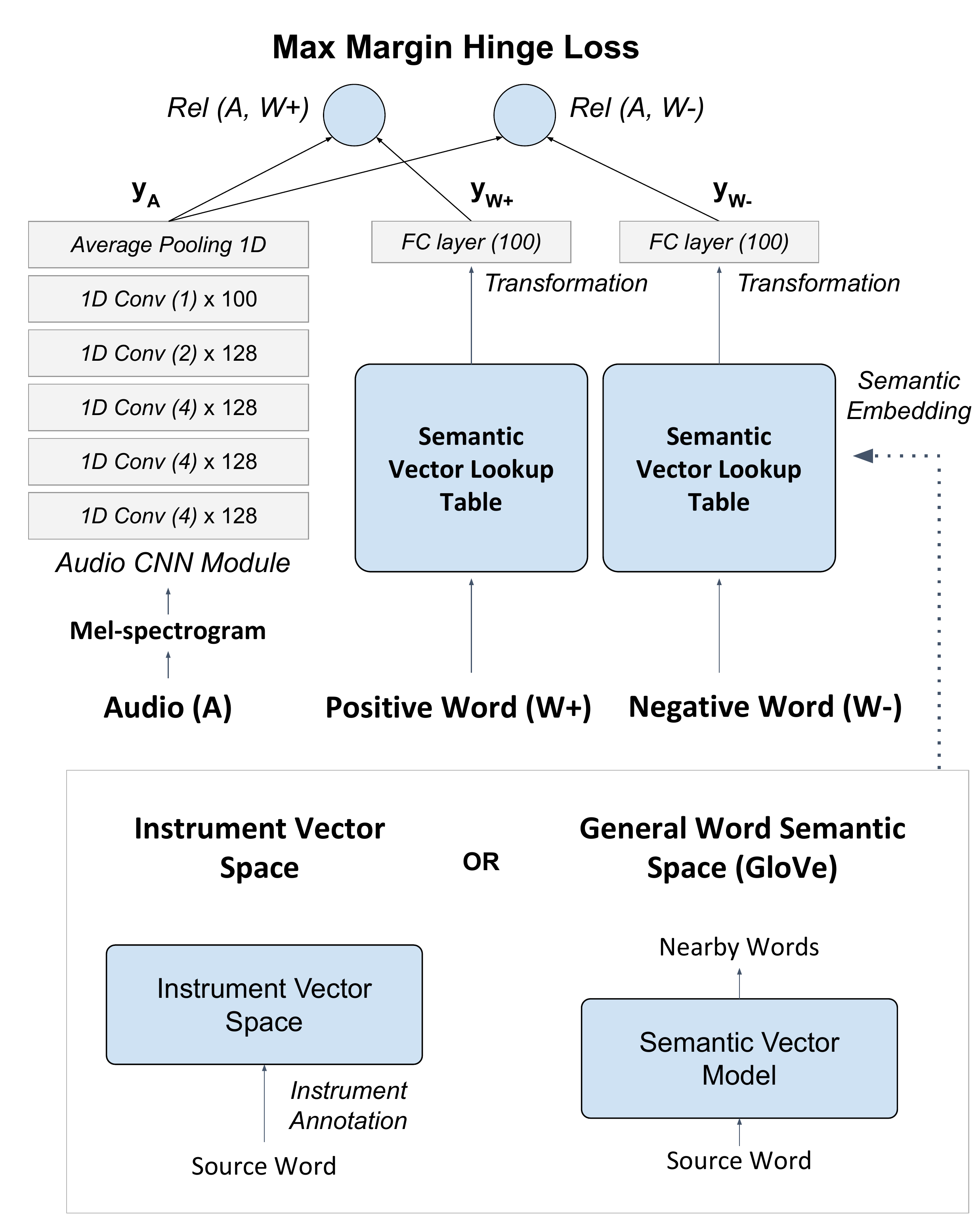}
\vspace{-3mm}
\caption{Deep embedding model for zero-shot learning using instrument vector space or general word semantic space.} 
\label{fig:figure3}
\end{figure}

Zero-shot learning is primarily performed by a compatibility function that maps multimedia embedding and semantic embedding. The joint embedding methods can be categorized into learning linear compatibility, nonlinear compatibility, intermediate attribute classifier, and their hybrid \cite{gzsl}. In this work, we focus on learning nonlinear compatibility. The model takes audio mel-spectrogram as input rather than audio embedding extracted from pre-trained model. A convolutional neural network (CNN) module for audio is learned directly with semantic embedding from ground truth annotations. Figure \ref{fig:figure3} illustrates the model architecture. The model takes audio from one module and randomly selected a positive word and a negative word from the ground truth annotations of the audio via the semantic vector lookup table. Following the previous works \cite{devise,park2017representation}, the loss function is chosen as a max-margin hinge loss as below:

\small
\[ \operatorname{L} (A,W) = \max \left[0, \Delta - Rel \left(A, W^{+} \right) + Rel \left(A,W^{-} \right) \right] \]
\normalsize
where $\Delta$ is the margin, $W^{+}$ denotes the label with positive annotation for the audio input, and $W^{-}$ denotes the label with negative annotation. The cosine similarity of the last hidden layer of the audio module and semantic module is used as a relevance score \cite{park2017representation}: 
\[ Rel(A, W) = Similarity_{cosine} \left(y_{A}, y_{W} \right) = \frac {y_{A}^{T} y_{W}} {|y_{A} \| y_{W}| }\]
where $y_{A}$ and $y_{W}$ denote the output of the last hidden layer for the audio module and the semantic module, respectively.

The mel-spectrogram based audio CNN module is constructed with four 1D convolutional layers with a 2D filter \cite{pons2016experimenting,dieleman2014end,choi2017convolutional,LeeNam2017}. Each layer is followed by a rectified linear unit (ReLU) activation and a max pooling layer. We added a convolutional layer and an average pooling layer on top of them to construct a fixed-size audio embedding vector compatible to the semantic module. The semantic module is constructed by adding a fully connected layer over a semantic embedding (the output of semantic vector lookup table). In this case, the semantic embedding (or semantic vector lookup table) can be composed with both human-annotated attributes data or general word semantic space. 

\subsection{Classification Model}
\label{sec:Classification_Model}
The deep embedding model can be used for not only the zero-shot learning setting but also the conventional multi-label classification and tagging problem where we train and evaluate the model with only seen labels. Therefore, we construct a classification model and compare it with the deep embedding model to evaluate the learning capability of the deep embedding model. The classification model is basically the same as the audio module of the deep embedding model but the output is the binary representation of the labels. To this end, we added a fully-connected output layer with the size of seen labels on top of the average pooling layer of the audio module. We used the sigmoid activation and binary cross-entropy loss for the multi-label classification.

\subsection{Training Details}
We extract mel-spectrogram from audio with 128 mel bins, 1024 size FFT (Hanning window), and 512 size hop in 22,050 Hz sampling rate. We standardized the mel-spectrogram across all training data to have zero mean and unit variance. We randomly selected three seconds of audio chunk (130 frames) as an input size to the CNN module. We optimized the model using stochastic gradient descent with 0.9 Nesterov momentum, 0.001 learning rate, and $1e^{-6}$ learning rate decay for all models and datasets. Our system is implemented in Python 3.5.2, Keras 2.2.2, and Tensorflow-gpu 1.6.0 for the back-end of Keras.\footnote{The source code is available at \url{https://github.com/kunimi00/ZSL_music_tagging}}

In the test phase, we took the average of the fixed-size audio embedding vectors over a single music track to obtain a track-level embedding, and made the predictions by the distance between track-level audio embedding and tag embedding.


\section{Experimental Settings}

We apply the proposed data split scheme and the deep embedding model to publicly available music classification and tagging datasets. We experiment with this in two settings of side information that we discussed in Section \ref{sec:section2}. For all the splits, we reserve 10\% instances of train set as a validation set randomly.\footnote{All the data splits are available along with the source code.}


\begin{table}[t]
\resizebox{1.0\columnwidth}{!}{
\begin{tabular}{@{}c ccccccc@{}}
\toprule
 & 
 \multicolumn{3}{c}{Label Split} & 
 \multicolumn{4}{c}{Instance Split} 
 \\ 
 \cmidrule(lr){2-4} 
 \cmidrule(l){5-8} 
 Dataset & 
 X (seen)        & 
 Y (unseen)      & 
 Total     & 
 A      & 
 B      & 
 C     & 
 Total  \\ 
 \midrule
FMA     & 125      & 32     & 157       & 11606   & 6935   & 925  & 19466  \\
MSD     & 900      & 226    & 1126      & 199385  & 188967 & 18057 & 406409    \\ 
\bottomrule
\end{tabular}
}
\caption{Data split statistics.}
\label{tab:table1}
\end{table}

\begin{table*}[t]
\centering
\resizebox{0.8\textwidth}{!}{\begin{tabular}{cccccccccccc}
\toprule
\multicolumn{2}{c}{Annotation Task} & \multicolumn{5}{c}{FMA} & \multicolumn{5}{c}{MSD} \\ \cmidrule(r){1-2} \cmidrule(lr){3-7} \cmidrule(l){8-12} 
Train & Test & AUC-i & MAP-i & P@1 & P@5 & P@10 & AUC-i & MAP-i & P@1 & P@5 & P@10 \\ \midrule
\multirow{6}{*}{A-X} 
 & B-Y & 0.8736 &	0.5118 &	0.3465 &	0.4810 &	0.5008 &	0.8559 &	0.3391 &	0.3895 &	0.2944 &	0.3109\\
 & C-Y &  0.7917 & 	0.3887 &	0.1563 &	0.0813 &	0.0643 &	0.8956 &	0.4536 &	0.3867 &	0.4278 &	0.4403 \\
 & (B+C)-Y &  0.8640 &	0.4973 &	0.3333 &	0.4648 &	0.4855 &	0.8593 &	0.3491 &	0.3893 &	0.3060 &	0.3221\\
 & B-(X+Y) & 0.8834 &	0.3524 &	0.3725 &	0.2604 &	0.3056 &	0.9107 &	0.2664 &	0.4653 &	0.2735 &	0.2265\\
 & C-(X+Y) &  0.7958 &	0.1316 &	0.0162 &	0.0841 &	0.1089 &	0.9019 &	0.1594 &	0.0522 &	0.1292 &	0.1449\\
 & (B+C)-(X+Y) & 0.9207 &	0.4264 &	0.4018 &	0.3540 &	0.3932 &	0.9315 &	0.2842 &	0.3762 &	0.2645 &	0.2504\\ \midrule
\multirow{6}{*}{B-X} 
 & B-Y & 0.8907 &	0.5633 &	0.4099 &	0.5371 &	0.5545 &	0.8679 &	0.3904 &	0.4803 &	0.3481 &	0.3623 \\
 & C-Y &  0.8144 &	0.4008 &	0.2281 &	0.3556 &	0.3834 &	0.8926 &	0.4814 &	0.4379 &	0.4573 &	0.4674 \\
 & (B+C)-Y & 0.8817 &	0.5442 &	0.3885 &	0.5158 &	0.5344 &	0.8700 &	0.3983 &	0.4766 &	0.3576 &	0.3715\\
 & B-(X+Y) & 0.9373	&   0.5142 &	0.5740 &	0.4206 &	0.4778 &	0.9217 &	0.2733 &	0.4462 &	0.2651 &	0.2283\\
 & C-(X+Y) & 0.8414 &	0.1886 &	0.0389 &	0.1494 &	0.1694 &	0.8941 &	0.1389 &	0.0666 &	0.1045 &	0.1231\\
 & (B+C)-(X+Y) & 0.8872 &	0.3606 &	0.3713 &	0.2881 &	0.3263 &	0.9276 &	0.2509 &	0.3272 &	0.2232 &	0.2147\\ \midrule
\multirow{6}{*}{(A+B)-X} 
 & B-Y & 0.8864 &	0.5036 &	0.3090 &	0.4748 &	0.4951 &	0.8632 &	0.3563 &	0.4349 &	0.3124 &	0.3278 \\
 & C-Y & 0.8286 &	0.4224 &	0.2530 &	0.3803 &	0.4064 &	0.8971 &	0.4836 &	0.4405 &	0.4591 &	0.4699 \\
 & (B+C)-Y &  0.8796 &	0.4940 &	0.3024 &	0.4636 &	0.4846 &	0.8662 &	0.3674 &	0.4354 &	0.3252 &	0.3402 \\
 & B-(X+Y) & 0.9118 &	0.4275 &	0.4784 &	0.3350 &	0.3813 &	0.9231 &	0.2977 &	0.5111 &	0.3094 &	0.2605\\
 & C-(X+Y) & 0.8276 &	0.1923 &	0.0530 &	0.1507 &	0.1738 &	0.9044 &	0.1920 &	0.1016 &	0.1629 &	0.1780 \\
 & (B+C)-(X+Y) &  0.9073 &	0.3812 &	0.3770 &	0.3064 &	0.3445 &	0.9370 &	0.2984 &	0.3987 &	0.2808 &	0.2660 \\ \bottomrule
\end{tabular}}
\vspace{-2mm}
\caption{Zero-Shot learning results for annotation task.}
\label{tab:table2}
\end{table*}

\subsection{Experiment 1: Genre with Instrument Attributes}
Musical instrument is one of the most important elements that determine music genre. Thus, recognizing instruments in a music track can be a strong cue to predict an unseen (or unheard) genre (assuming that the predictor learned the genre from some literature). For this experiment, we use Free Music Archive (FMA) \cite{fma} and OpenMIC-2018 datasets \cite{openmic}. FMA contains audio files and genre annotations. OpenMIC-2018, which was originally designed for multiple instrument recognition, has 20 different instrument annotations to the audio files in FMA. We filtered the audio files to have both genre and instrument annotations. As a result, 19,466 audio with 157 genre labels and 20 instrument annotations are left. Following the proposed data split scheme, we randomly split 157 labels into 125 seen and 32 unseen ones. Then, three groups of audio instances (A, B, C) are created naturally. The statistics of audio and labels is described in Table \ref{tab:table1}.

The annotations on 20 different instruments in the OpenMIC-2018 dataset are labeled as a likelihood measure summarized from crowd-sourced annotations. They are regarded as positive if the likelihood value of an instrument annotation is larger than 0.5 and, otherwise, negative. The exact value of 0.5 means that it's unannotated. So, we treated positive and negative information independently and created 40 dimensional instrument vectors that can represent both positive and negative information. The genre to instrument attribute relationship is then constructed by accumulating 40 dimensional instrument vectors of the songs according to the genre labels. Finally, we standardized the instrument vectors to have zero mean and unit variance. This was used as the semantic vector lookup table in the learning model.


\subsection{Experiment 2: General Word Semantic Space}

The other type of side information is word embedding learned from a large-scale text dataset separately from music datasets. It can represent words as vectors in a semantic space. We adopted GloVe as a word embedding model \cite{glove}. Instead training it from scratch, we used a publicly available pre-trained GloVe model\footnote{The Common Crawl model was trained with 42B tokens containing 1.9M vocabulary. https://nlp.stanford.edu/projects/glove/}. It consists of 300-dimensional vectors of 19 million vocabularies trained from documents in Common Crawl data. Since this can cover a large vocabulary of words, we used a music dataset with rich annotations, which is Million Song Dataset (MSD) with the Last.fm tag annotations \cite{msd}. From the full set of 498,035 tags in the Last.fm annotations, 
we filtered the tags that correspond to 2,000 genre/sub-genre classes contained in Tagtraum genre ontology \cite{tagtraum}. We filtered the result (1800 tags) again into 1,126 tags after eliminating missing words in the dictionary of the pre-trained GloVe model. We used 406,409 audio instances annotated with the refined 1,126 tags. Following the proposed data split scheme, we randomly split 1,126 tags into 900 seen and 226 unseen ones, and organized three groups of audio instances (A, B and C). They are summarized in Table \ref{tab:table1}.

\subsection{Evaluation Metrics}
We used the area under the ROC curve averaged over instance (AUC-i),  mean average precision over instance (MAP-i), and precision at K (P@K) as evaluation metrics for the annotation task. The retrieval task is evaluated using the area under the ROC curve averaged over label (AUC-l) and mean average precision over label (MAP-l).

\begin{table}[t]
\centering
\resizebox{0.85\columnwidth}{!}{\begin{tabular}{@{}cccccc@{}}
\toprule
\multicolumn{2}{c}{Data Split} & \multicolumn{2}{c}{FMA} & \multicolumn{2}{c}{MSD} \\ \cmidrule(r){1-2} \cmidrule(lr){3-4} \cmidrule(l){5-6}
Train & Test & AUC-l & MAP-l & AUC-l & MAP-l \\ \midrule
\multirow{2}{*}{A-X} 
 & (B+C)-Y & 0.6793 &	0.0904 & 0.6740	& 0.0279   \\
 & (A+B+C)-Y & 0.6771 &	0.0392 & 0.6673	& 0.0149   \\ \midrule
\multirow{2}{*}{B-X} 
 & (B+C)-Y & 0.7194 &	0.1280  &  0.6907 &	0.0295  \\
 & (A+B+C)-Y & 0.7236 &	0.0662  & 0.6843 &	0.0158  \\ \midrule
\multirow{2}{*}{(A+B)-X} 
 & (B+C)-Y & 0.7314 &	0.1170   & 0.6864 &	0.0310    \\
 & (A+B+C)-Y & 0.7377 &	0.0518 & 0.6789 &	0.0172   \\ \bottomrule
\end{tabular}}
\caption{Zero-Shot learning results for retrieval task.}
\label{tab:table3}
\end{table}

\section{Results}

\subsection{Multi-label Zero-Shot Annotation}
We compare the results of the combination of the proposed multi-label zero-shot learning split in the annotation task. 
From Table \ref{tab:table2}, we can find that training with (A+B) or B instance set shows better performance than that with A in general. This indicates that the instances in B give better supervision over the entire tag set. In the case of test on C-(X+Y), the MAP-i and P@K scores are very low. This is because the case is generalized zero-shot learning evaluation setting \cite{gzsl} which makes predictions of seen label even when the ground truth of seen labels have only negatives.

We also see that some results have different trends between the two datasets. For example, test on B gives better results than C on FMA, but the results are opposite to those on MSD.  
We suspect that this may be due to difference in label cardinality, the average number of labels per instance \cite{multioverview}, which can significantly affect performance in multi-label classification \cite{cardinality}. Specifically, FMA tracks have cardinality of 1.18 for B-Y and 1.15 for C-Y, whereas MSD tracks have cardinality of 2.04 in B-Y and only 1.11 in C-Y. This lower cardinality may cause better performance in C-Y than B-Y for MSD. However, we need further investigation considering differences in datasets and side information. 


\begin{table}[t]
\centering
\resizebox{0.85\columnwidth}{!}{\begin{tabular}{@{}ccccc@{}}
\toprule
\multicolumn{3}{c}{Data Split} & FMA & MSD \\ 
\cmidrule(r){1-3} \cmidrule(lr){4-4} \cmidrule(l){5-5}
Train & Test & Model & AUC-l & AUC-l \\ \midrule
\multirow{2}{*}{A-X} & \multirow{2}{*}{B-X} & Deep Embedding Model & 0.7381 & 0.7161 \\
 &  & Classification Model & 0.7250 &  0.6980 \\ 
 \bottomrule
\end{tabular}}
\caption{Results on retrieval task that compares deep embedding model to classification model. The detail of classification model is described in Section \ref{sec:Classification_Model}.}
\label{tab:table4}

\end{table}

\subsection{Multi-label Zero-Shot Retrieval}
The results of the retrieval task are reported in Table \ref{tab:table3}. As in the annotation task, the overall performance is high when training with (A+B)-X. Also, the test results on (A+B+C)-Y are lower than that on (B+C)-Y. We can regard the test on (A+B+C)-Y as a generalized zero-shot learning evaluation setting for the retrieval task. This means that even for instances that do not have a positive annotation on unseen label according to the split (instances that were denoted as A), the evaluation is performed including this instances so to consider whether the model actually make a negative prediction on them. Thus, this is a more strict evaluation setting.

\subsection{Deep Embedding Model vs. Classification Model}

We also conducted an additional experiment to verify the performance of the deep embedding model in the conventional  multi-label  classification  and  tagging  task\footnote{In this evaluation, some labels were excluded because there are some labels that have negative annotations for all instances according to the split.}. Table \ref{tab:table4} shows results in the retrieval scenario following previous work \cite{pons2016experimenting,dieleman2014end,choi2017convolutional,LeeNam2017}. We can see that the deep embedding model outperforms the classification model on both datasets. This indicates that associating audio with labels via the side information is a powerful approach even in the conventional multi-label classification and tagging  task.

\begin{table}[!t]
\centering
\resizebox{1.0\columnwidth}{!}{\begin{tabular}{@{}ccc@{}}
\toprule
\multirow{2}{*}{Query}
& Top 5 Retrieved Tracks
& Original Last.fm  
\\ 
& (Title / Artist)                                   & Annotation 
\\ 
\midrule
\multirow{10}{*}{guitar}                    & Iron Acton /                             &  psychedelic, experimental,
\\
     & Beak    & krautrock, english, bass                       
\\
\cmidrule(l){2-3}
     & Drink Whiskey And Shut up / & rock                                   
\\
    & Brian Setzer & 
\\
\cmidrule(l){2-3}
     & Thar She Blows /          
& party, surf, 
\\
      & The Halibuts                              & surfrock
\\
\cmidrule(l){2-3}
     & All Quiet On 34th Street / &  rock,  rocknroll,  hardrock, \\
      & Eric Burdon And The New Animals &  screamo,  00s 
\\
\cmidrule(l){2-3}
     & Gimme Some Lovin' /                                 & 60s, british, 
\\ 
      &  Traffic                                & classicrock, rock
\\
\midrule
\multirow{10}{*}{lovely}   
& Eddie My Love / 
& 50s, doowop, oldies, 
\\
& The Chordettes                             
& pop, vocal
\\
\cmidrule(l){2-3}
& Vaya Con Dios / 
& 50s, jazz, 
\\
& Les Paul \& Mary Ford                      
& oldies
\\
\cmidrule(l){2-3}
& (I Can't Help You) I'm Falling Too /         
& country, 
\\
& Skeeter Davis         
& oldies
\\
\cmidrule(l){2-3}
& Mr. Blue / 
& oldies, 50s, pop, 
\\
& The Fleetwoods
& doowop, ballad
\\
\cmidrule(l){2-3}
& I'm Blue Again / 
& blue, 
\\
& Patsy Cline
& country
\\ \bottomrule
\end{tabular}
}
\caption{Top 5 retrieved tracks for a query word from unseen tag subset (`guitar') and an arbitrary word (`lovely').}
\label{tab:ret_song}

\end{table}

\begin{table}[t]
\centering
\resizebox{1\columnwidth}{!}{\begin{tabular}{@{}ccc@{}}
\toprule
\begin{tabular}[c]{@{}c@{}}Smells like teen spirit\\ Nirvana\end{tabular} & \begin{tabular}[c]{@{}c@{}}Superstition\\ Stevie Wonder\end{tabular} & \begin{tabular}[c]{@{}c@{}}Theme to Grace / Lament \\ George Winston\end{tabular} \\ 
\midrule
classicrock (unseen) & funk (unseen)  & folk                                                     \\
punk                                                            & soul                                                        & instrumental                                                     \\
rock (unseen)                                                          & pop (unseen)         & jazz                                               \\
80s (unseen)        & jazz                                                        & piano (unseen)                                                \\
alternative                                                     & 80s (unseen)                                         & singersongwriter                                                       \\
punkrock                                                         & blues (unseen)          & chillout                                                              \\
90s                      & classicrock                                                & acoustic                                                      \\
metal                                                             & 90s (unseen)                             & blues       \\
vintage                                                          & disco                                                      & mellow                                                    \\
alternativerock                          & dance                                                      & chill                                 \\
\bottomrule
\end{tabular}}
\caption{Top 10 auto-tagging results for examples of well-known songs including unseen tags during training.}
\label{tab:annotation}
\end{table}

\begin{table}[t]
\centering
\resizebox{1\columnwidth}{!}{\begin{tabular}{@{}ccc@{}}
\toprule
Query
& General Semantic Space                                    
& \begin{tabular}[c]{@{}c@{}}
Zero-shot Embedding \\ Space 
\end{tabular}
\\ 
\midrule
guitar
& 
\begin{tabular}[c]{@{}c@{}}
bass, acoustic, piano, vocals, \\ 
violin, percussion, strings, \\
vocal, music, jazz
\end{tabular}
  & 
\begin{tabular}[c]{@{}c@{}}  
 instrumental, minimal, rock, \\
 acidrock, progressiverock, \\
 alternative, psychedelic, \\ 
 folkrock, classicrock, band  
\end{tabular}
\\
\midrule
lovely
& 
\begin{tabular}[c]{@{}c@{}}
awesome, love, cool, romantic,\\ 
relaxing, summer, christmas, \\
holiday, vintage, soft
\end{tabular}
  & 
\begin{tabular}[c]{@{}c@{}}  
relaxing, relax, lovesongs, \\
easylistening, baby, country, \\
romantic, easy, americana, \\
ballad  
\end{tabular}
\\
\bottomrule
\end{tabular}}
\caption{Comparison of top 10 nearest word vectors (out of 1,126 tags) on general semantic space and the trained zero-shot embedding space.}
\label{tab:ret_word}
\end{table}

\subsection{Case Study}
We conducted case studies to better understand the performance of the zero-shot learning model. Table \ref{tab:annotation} shows the results of annotation on several famous music tracks. They show that the predictions are reasonable for both seen or unseen tags. Table \ref{tab:ret_song} lists retrieved tracks given a query word and their original labels. We can see that the results are reasonable even if we did not use seen tags. 
Furthermore, we compared the general semantic space of GloVe model to our trained deep embedding space in Table \ref{tab:ret_word}. The examples show the deep embedding space learns the relationship between words in a more musically meaningful way.


\section{Conclusions}
In this paper, we showed that zero-shot learning is capable of associating music audio with unseen labels using side information. This allow to use a rich vocabulary of words to describe music, thereby enhancing the experience of music retrieval or recommendation in a more human-friendly way. There is a large room to explore for future work. For example, lyrics can be used as side information which can be obtained without manual human annotation. The neural language models can be also trained to contain more musical context, for example, using text descriptions of playlists or music articles. 




\section{Acknowledgment}
This work was supported by NAVER Corp.

\bibliography{ISMIRtemplate}

\begin{thebibliography}{10}

\bibitem{akata2015SJE}
Zeynep Akata, Scott Reed, Daniel Walter, Honglak Lee, and Bernt Schiele.
\newblock Evaluation of output embeddings for fine-grained image
  classification.
\newblock In {\em Proc. of The IEEE Conference on Computer Vision and Pattern
  Recognition (CVPR)}, pages 2927--2936, 2015.

\bibitem{al2016recovering}
Ziad Al-Halah, Makarand Tapaswi, and Rainer Stiefelhagen.
\newblock Recovering the missing link: Predicting class-attribute associations
  for unsupervised zero-shot learning.
\newblock In {\em Proc. of The IEEE Conference on Computer Vision and Pattern
  Recognition (CVPR)}, pages 5975--5984, 2016.

\bibitem{msd}
Thierry Bertin-Mahieux, Daniel~P.W. Ellis, Brian Whitman, and Paul Lamere.
\newblock The million song dataset.
\newblock In {\em Proc. of International Society for Music Information
  Retrieval Conference (ISMIR)}, 2011.

\bibitem{cardinality}
Francisco Charte, Antonio Rivera, Mar{\'\i}a~Jos{\'e} del Jesus, and Francisco
  Herrera.
\newblock Improving multi-label classifiers via label reduction with
  association rules.
\newblock In {\em International Conference on Hybrid Artificial Intelligence
  Systems}, pages 188--199. Springer, 2012.

\bibitem{choi2017convolutional}
Keunwoo Choi, Gy{\"o}rgy Fazekas, Mark Sandler, and Kyunghyun Cho.
\newblock Convolutional recurrent neural networks for music classification.
\newblock In {\em Proc. of the IEEE International Conference on Acoustics,
  Speech, and Signal Processing (ICASSP)}, pages 2392--2396, 2017.

\bibitem{fma}
Micha\"el Defferrard, Kirell Benzi, Pierre Vandergheynst, and Xavier Bresson.
\newblock Fma: A dataset for music analysis.
\newblock In {\em Proc. of International Society for Music Information
  Retrieval Conference (ISMIR)}, 2017.

\bibitem{dieleman2014end}
Sander Dieleman and Benjamin Schrauwen.
\newblock End-to-end learning for music audio.
\newblock In {\em Proc. of the IEEE International Conference on Acoustics,
  Speech and Signal Processing (ICASSP)}, pages 6964--6968, 2014.

\bibitem{devise}
Andrea Frome, Greg~S Corrado, Jon Shlens, Samy Bengio, Jeff Dean, Marc~Aurelio
  Ranzato, and Tomas Mikolov.
\newblock Devise: A deep visual-semantic embedding model.
\newblock In C.~J.~C. Burges, L.~Bottou, M.~Welling, Z.~Ghahramani, and K.~Q.
  Weinberger, editors, {\em Advances in Neural Information Processing Systems
  26}, pages 2121--2129. Curran Associates, Inc., 2013.

\bibitem{fu2017recent}
Yanwei Fu, Tao Xiang, Yu-Gang Jiang, Xiangyang Xue, Leonid Sigal, and Shaogang
  Gong.
\newblock Recent advances in zero-shot recognition: Toward data-efficient
  understanding of visual content.
\newblock {\em IEEE Signal Processing Magazine}, 35(1):112--125, 2018.

\bibitem{fu2015transductive}
Yanwei Fu, Yongxin Yang, Tim Hospedales, Tao Xiang, and Shaogang Gong.
\newblock Transductive multi-label zero-shot learning.
\newblock {\em British Machine Vision Association}, pages 7.1--7.11, 2014.

\bibitem{openmic}
Eric Humphrey, Simon Durand, and Brian McFee.
\newblock Openmic-2018: an open dataset for multiple instrument recognition.
\newblock In {\em Proc. of International Society for Music Information
  Retrieval Conference (ISMIR)}, 2018.

\bibitem{awa}
Christoph~H Lampert, Hannes Nickisch, and Stefan Harmeling.
\newblock Attribute-based classification for zero-shot visual object
  categorization.
\newblock {\em IEEE Transactions on Pattern Analysis and Machine Intelligence},
  36(3):453--465, 2014.

\bibitem{LeeNam2017}
Jongpil Lee and Juhan Nam.
\newblock Multi-level and multi-scale feature aggregation using pretrained
  convolutional neural networks for music auto-tagging.
\newblock {\em IEEE Signal Processing Letters}, 24(8):1208--1212, 2017.

\bibitem{costa}
Thomas Mensink, Efstratios Gavves, and Cees~GM Snoek.
\newblock Costa: Co-occurrence statistics for zero-shot classification.
\newblock In {\em Proc. of The IEEE Conference on Computer Vision and Pattern
  Recognition (CVPR)}, pages 2441--2448, 2014.

\bibitem{skipgram}
Tomas Mikolov, Kai Chen, Gregory~S. Corrado, and Jeffrey Dean.
\newblock Efficient estimation of word representations in vector space.
\newblock In {\em Proc. of International conference on Learning Representations
  (ICLR)}, 2013.

\bibitem{word2vec}
Tomas Mikolov, Ilya Sutskever, Kai Chen, Greg~S Corrado, and Jeff Dean.
\newblock Distributed representations of words and phrases and their
  compositionality.
\newblock In {\em Advances in neural information processing systems}, pages
  3111--3119, 2013.

\bibitem{wordnet}
George~A Miller.
\newblock Wordnet: a lexical database for english.
\newblock {\em Communications of the ACM}, 38(11):39--41, 1995.

\bibitem{nam2019}
Juhan Nam, Keunwoo Choi, Jongpil Lee, Szu-Yu Chou, and Yi-Hsuan Yang.
\newblock Deep learning for audio-based music classification and tagging:
  Teaching computers to distinguish rock from bach.
\newblock {\em IEEE Signal Processing Magazine}, pages 41--51, 2019.

\bibitem{conse}
Mohammad Norouzi, Tomas Mikolov, Samy Bengio, Yoram Singer, Jonathon Shlens,
  Andrea Frome, Greg~S Corrado, and Jeffrey Dean.
\newblock Zero-shot learning by convex combination of semantic embeddings.
\newblock In {\em Proc. of International conference on Learning Representations
  (ICLR)}, 2013.

\bibitem{palatucci2009zero}
Mark Palatucci, Dean Pomerleau, Geoffrey~E Hinton, and Tom~M Mitchell.
\newblock Zero-shot learning with semantic output codes.
\newblock In {\em Advances in neural information processing systems}, pages
  1410--1418, 2009.

\bibitem{parikh2011relative}
Devi Parikh and Kristen Grauman.
\newblock Relative attributes.
\newblock In {\em International Conference on Computer Vision (ICCV)}, pages
  503--510. IEEE, 2011.

\bibitem{park2017representation}
Jiyoung Park, Jongpil Lee, Jangyeon Park, Jung{-}Woo Ha, and Juhan Nam.
\newblock Representation learning of music using artist labels.
\newblock In {\em Proc. of International Society for Music Information
  Retrieval Conference (ISMIR)}, pages 717--724, 2018.

\bibitem{glove}
Jeffrey Pennington, Richard Socher, and Christopher Manning.
\newblock Glove: Global vectors for word representation.
\newblock In {\em Proc. of the 2014 conference on empirical methods in natural
  language processing (EMNLP)}, pages 1532--1543, 2014.

\bibitem{pons2016experimenting}
Jordi Pons, Thomas Lidy, and Xavier Serra.
\newblock Experimenting with musically motivated convolutional neural networks.
\newblock In {\em Proc. of the International Workshop on Content-Based
  Multimedia Indexing (CBMI)}, pages 1--6, 2016.

\bibitem{reed2016learning}
Scott Reed, Zeynep Akata, Honglak Lee, and Bernt Schiele.
\newblock Learning deep representations of fine-grained visual descriptions.
\newblock In {\em Proc. of The IEEE Conference on Computer Vision and Pattern
  Recognition (CVPR)}, pages 49--58, 2016.

\bibitem{ren2015multi}
Zhou Ren, Hailin Jin, Zhe Lin, Chen Fang, and Alan Yuille.
\newblock Multiple instance visual-semantic embedding.
\newblock In {\em Proc. of the British Machine Vision Conference (BMVC)},
  volume~1, page~3, 2017.

\bibitem{rohrbach2010whathelps}
Marcus Rohrbach, Michael Stark, Gy{\"o}rgy Szarvas, Iryna Gurevych, and Bernt
  Schiele.
\newblock What helps where--and why? semantic relatedness for knowledge
  transfer.
\newblock In {\em Proc. of The IEEE Conference on Computer Vision and Pattern
  Recognition (CVPR)}, pages 910--917. IEEE, 2010.

\bibitem{sandouk2016multi}
Ubai Sandouk and Ke~Chen.
\newblock Multi-label zero-shot learning via concept embedding.
\newblock {\em arXiv preprint arXiv:1606.00282}, 2016.

\bibitem{tagtraum}
Hendrik Schreiber.
\newblock Genre ontology learning: Comparing curated with crowd-sourced
  ontologies.
\newblock In {\em Proc. of International Society for Music Information
  Retrieval Conference (ISMIR)}, pages 400--406, 2016.

\bibitem{singh2016end}
Krishna~Kumar Singh and Yong~Jae Lee.
\newblock End-to-end localization and ranking for relative attributes.
\newblock In {\em European Conference on Computer Vision}, pages 753--769.
  Springer, 2016.

\bibitem{multioverview}
Grigorios Tsoumakas and Ioannis Katakis.
\newblock Multi-label classification: An overview.
\newblock {\em International Journal of Data Warehousing and Mining (IJDWM)},
  3(3):1--13, 2007.

\bibitem{turnbull2008semantic}
Douglas Turnbull, Luke Barrington, David Torres, and Gert Lanckriet.
\newblock Semantic annotation and retrieval of music and sound effects.
\newblock {\em IEEE/ACM Transactions on Audio, Speech, and Language
  Processing}, 16(2):467--476, 2008.

\bibitem{wang2017multi}
Qian Wang and Ke~Chen.
\newblock Multi-label zero-shot human action recognition via joint latent
  embedding.
\newblock {\em arXiv preprint arXiv:1709.05107}, 2017.

\bibitem{wang2019survey}
Wei Wang, Vincent~W Zheng, Han Yu, and Chunyan Miao.
\newblock A survey of zero-shot learning: Settings, methods, and applications.
\newblock {\em ACM Transactions on Intelligent Systems and Technology (TIST)},
  10(2):13, 2019.

\bibitem{welinder2010caltech}
Peter Welinder, Steve Branson, Takeshi Mita, Catherine Wah, Florian Schroff,
  Serge Belongie, and Pietro Perona.
\newblock Caltech-{UCSD} birds 200.
\newblock 2010.

\bibitem{xian2016latent}
Yongqin Xian, Zeynep Akata, Gaurav Sharma, Quynh Nguyen, Matthias Hein, and
  Bernt Schiele.
\newblock Latent embeddings for zero-shot classification.
\newblock In {\em Proc. of The IEEE Conference on Computer Vision and Pattern
  Recognition (CVPR)}, pages 69--77, 2016.

\bibitem{gzsl}
Yongqin Xian, Bernt Schiele, and Zeynep Akata.
\newblock Zero-shot learning - the good, the bad and the ugly.
\newblock In {\em Proc. of The IEEE Conference on Computer Vision and Pattern
  Recognition (CVPR)}, July 2017.

\end{thebibliography}

%
%
%
%

\end{document}